\documentclass{llncs}
\usepackage{tabularx}
\usepackage{graphicx}
\usepackage{multirow}
\usepackage{enumitem}

\usepackage{amsmath,amssymb,amsfonts}
\usepackage{amsthm}
\usepackage{mathrsfs}
\usepackage[title]{appendix}
\usepackage{xcolor}
\usepackage{textcomp}
\usepackage{manyfoot}
\usepackage{booktabs}
\usepackage{algorithm}
\usepackage{algorithmicx}
\usepackage{algpseudocode}
\usepackage[most]{tcolorbox}
\usepackage{adjustbox}
\usepackage{subcaption}
\usepackage{xurl}
\usepackage{hyperref}

\begin{document}

\title{Analysing User Reviews to Identify User Concerns Around Permissions in AI Apps}


\author{
Babar Shah\inst{1} \and
Faheem Ullah\inst{2} \and
Myles Watkinson\inst{3} \and
Muhammad Moiz Khalid\inst{4} \and
Tehmina Karamat Khan\inst{5} \and
Muhammad Junaid\inst{6}
}

\institute{
College of Technological Innovation, Zayed University, UAE\\
\email{babar.shah@zu.ac.ae}
\and
College of Interdisciplinary Studies, Zayed University, UAE\\
\email{faheem.ullah@zu.ac.ae}
\and
University of Adelaide, Adelaide, South Australia, Australia\\
\email{myleswatkinson1@gmail.com}
\and
National University of Computer and Emerging Sciences, Pakistan\\
\email{i232552@isb.nu.edu.pk}
\and
School of Computer Science, Taylor's University, Malaysia\\
\email{Tehmina.Khan@taylors.edu.my}
\and
Dept of Statistics, Islamia College University, Peshawar, Pakistan\\
\email{m.junaid7886@gmail.com}
}

\maketitle

\begin{abstract}
Artificial intelligence is increasingly embedded in everyday software, making its integration into mobile apps inevitable. However, AI mobile app developers are not always versed in security and privacy best practices, leaving users to monitor their own security and understand how apps use their data. App reviews capture real user experiences, helping others make informed decisions before downloading. This paper presents a machine learning model for classifying AI app reviews into permission-related categories. Because user reviews are unstructured, assembling a conventional labeled training set is difficult. To address this, AI-generated security and permission reviews are used to identify relevant training examples from a large corpus of human-written reviews, eliminating the need for manual annotation. The proposed approach classified permission reviews with an accuracy of 82\%. Analysis shows that users organise their concerns by sentiment toward the requesting app rather than specific permission types, with implications for users, developers, and platform administrators.

\end{abstract}

\section{Introduction}
Mobile app stores such as the iOS App Store and Google Play Store let users leave feedback on the apps they've downloaded \cite{apple_reviews_2019}. Unlike structured surveys or controlled studies, these reviews reflect genuine user experiences, and one dimension users write about with particular frequency is security and privacy. Privacy has become a major global concern and previous studies show that a large number of app reviews discuss security related concerns including phishing, data leakage, and permission misuse \cite{nema2022,tao2020,F1}. Permissions sit at the heart of the privacy relationship between a user and an app: before accessing sensitive data such as a user's location, contacts, or microphone, an app must request the user's explicit consent \cite{apple_permissions}, allowing users to evaluate whether an app should have access, which can influence others' download decisions. Mobile apps can request up to 235 distinct permissions, far more than most users would be aware of, raising the concern that users routinely share more data than they intend to \cite{olmstead2015}. This concern is amplified in AI mobile apps, which typically require access to user data to train and refine their models, and as Andreotta et al. note, meaningful transparency and user control over data in AI systems remain elusive \cite{andreotta2021}.

Despite growing work on security review classification, existing research has not fully addressed this problem. Prior studies have demonstrated that security-related reviews can be identified from large datasets using NLP-based classifiers \cite{nema2022,tao2020,nguyen2019}, but these efforts stop at classification without examining what the review says or why the user wrote it, and focus only on app stores rather than AI-specific apps, leaving open whether users of AI apps exhibit distinct patterns of concern.

This study asks whether users are concerned about all permission requests or only specific types. AI-generated security and permission reviews are used to select relevant training examples from human-written reviews, offering a semantically grounded alternative to n-gram selection, training binary neural network classifiers. Once permission-related reviews are identified, K-means clustering surfaces the topics discussed, compared across app genres. The first contribution is a classifier for permission related reviews with an accuracy of 82\%; the next is an analysis of topics discussed, leading to the following research questions:

\begin{itemize}[label=\textbullet]
\item \textit{RQ1: Can permission-related reviews be classified from a dataset of user reviews?}
\item \textit{RQ2: What topics do users discuss in permission-related reviews?}
\end{itemize}

The rest of this paper is organized as follows. Section \ref{lit} reviews the relevant literature. Section \ref{method} describes the methodology and section \ref{results} presents the results of the classification and clustering experiments. Section \ref{dis} discusses the findings and section \ref{limit} reflects on the limitations of the study. Section \ref{conc} concludes the paper.

\section{Literature Review}\label{lit}
This section lays out relevant literature across three interconnected areas: user attitudes towards permission requests, security issues in AI mobile apps, and large-scale analysis of user reviews.

\subsection{User Attitudes towards App Permissions}
Felt et al. found that 70\% of users considered reviews important for judging an app's security, yet only half had a clear understanding of the app's actual permissions \cite{felt_permissions}, which, paired with Olmstead and Atkinson's finding that apps can request up to 235 distinct permissions, suggests users may expose far more data than they realize \cite{olmstead2015}.

\subsection{Security Issues in AI mobile Apps}
Pistoia, et al. applied ML techniques to flag high-risk code in open source apps \cite{pistoia2017}, while Chatterjee, et al. found nearly a quarter of mobile apps contain issues with direct consequences for users \cite{chatterjee2016}. Hu, et al. identified security threats at each stage of the AI lifecycle \cite{hu2021}, and Andreotta, et al. found users frequently lack meaningful transparency or control over their data \cite{andreotta2021}. Machine learning approaches have similarly been applied to detect security threats such as data exfiltration \cite{F2}. Large-scale cybersecurity analytics systems have also been examined to identify architectural strategies and adaptive designs for handling such threats \cite{F3,F4}. This bleeds into our research, as Mukherjee, et al. found that permissions surrounding data usage was the most common topic in user reviews of mobile apps \cite{mukherjee2020}.

\subsection{User Reviews Analysis}
Panichella, et al. and Chen, et al. \cite{panichella2015} \cite{chen2014} classified user reviews using NLP to analyse structure, semantics, and sentiment, while Palomba, et al. clustered reviews to rank topics for developers \cite{palomba2017}. Most closely related to this paper, Nema, et al., Tao, et al. and Nguyen, et al. \cite{nema2022} \cite{tao2020} \cite{nguyen2019} each train binary classifiers to extract security or privacy related reviews. Nguyen, et al. used a bag of words (BOW) approach, though BOW is susceptible to false positives since keyword matching cannot distinguish opposing sentiments toward the same term \cite{nema2022} \cite{tao2020}. Nema, et al. instead used BERT and USE \cite{devlin2019} \cite{cer2018} for semantic similarity, achieving a precision of around 0.9 using security n-grams for training data selection \cite{nema2022}, motivated by data scarcity \cite{lu2021}, though generated data's quality and bias must be assessed before use \cite{rajotte2022}.

This paper builds on this work: existing classifiers only extract reviews on specific topics, whereas the present study first classifies security reviews, then isolates permission-related reviews within that set, investigating which permission types concern users most across genres \cite{grano2017}.

\section{Methodology}\label{method}
\begin{figure}[htbp!]
\centering
\includegraphics[width=\textwidth]{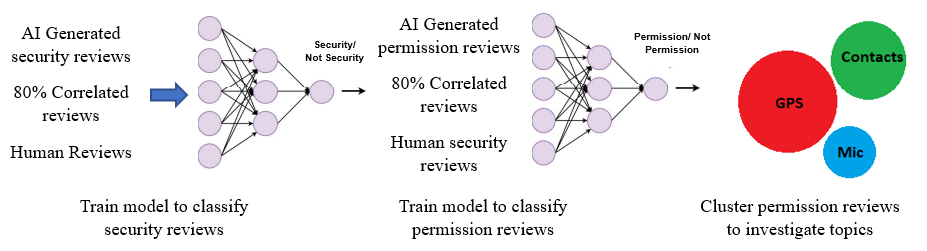}
\caption{Methodology pipeline.}
\label{fig:methodology}
\end{figure}

The project execution consisted of multiple stages: data curation, selection of training and test data, ML model training and data analysis. As this study aims to classify privacy-related reviews and further identify permission-related reviews, model training was conducted twice. The complete methodology is illustrated in Figure \ref{fig:methodology}.

\subsection{Data Curation}
\begin{figure}[htbp!]
\centering
\begin{subfigure}[b]{0.48\linewidth}
\centering
\includegraphics[width=\linewidth]{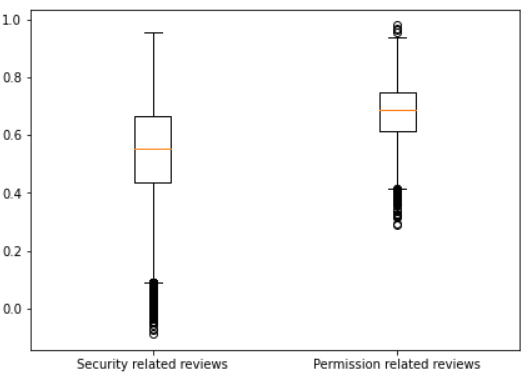}
\caption{Similarity matrix values.}
\label{fig:similarity}
\end{subfigure}
\hfill
\begin{subfigure}[b]{0.51\linewidth}
\centering
\includegraphics[width=\linewidth]{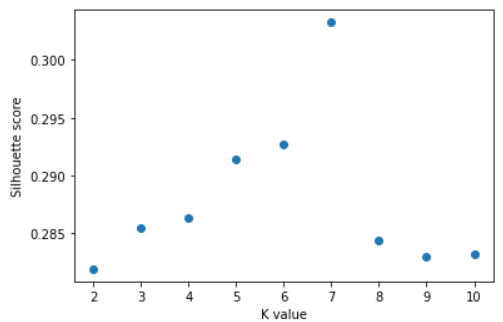}
\caption{Silhouette score by k.}
\label{fig:silhouette}
\end{subfigure}
\caption{Similarity matrix box plots (left) and silhouette score across k values (right), used respectively to assess review diversity and select the optimal number of clusters.}
\label{fig:similarity-silhouette}
\end{figure}

Two types of data were collected: human-generated reviews (HR) and AI-generated reviews (GR). HRs were scraped using google\_play\_scraper \cite{eason2021} from Li et al.'s dataset of 56,682 AI apps \cite{li2022}, yielding 16,142,572 HRs. GRs were collected using OpenAI's GPT-4 \cite{openai2023}, with prompts varying semantics, subject matter, and sentiment to address data scarcity (Section 2.3), producing 550 security reviews and 100 permission reviews. Each GR was encoded with USE \cite{cer2018} to assess uniqueness via similarity matrices (Figure~\ref{fig:similarity}), with median similarity scores of 0.5 (security) and 0.6 (permission) providing enough variability for reliable identification of related reviews.

\vspace{-25pt}
\vspace{-20pt}
For converting these reviews to usable data, pre-processing steps were performed including correcting typographical errors, eliminating duplicate words, tokens and punctuation, and removal of stop words and emojis.

\subsection{Selection of training and test data}
Training a neural network reliably requires data 10 times larger than the number of output features \cite{cho2016}; since USE encodes each sentence into a 512-dimensional vector, at least 5,120 reviews were needed, exceeding those available. Hence, GRs were used to identify additional examples from the HR datasets, improving on the n-gram based selection used in prior work \cite{nema2022}: GRs were converted into vector representations and each HR compared against them, with any HR correlating above 80\% treated as relating to the same topic. This yielded 6,187 security-related HRs, matched with an equal number of randomly selected reviews labeled `not security related', with a manually annotated test set of 50 security-related and 50 unrelated reviews. The process was repeated for permission reviews using permission GRs.

\subsection{Model Training}
Two binary classifiers were developed to identify privacy-related reviews from the full dataset and then to distinguish permission-related reviews from within the security review subset. The permission classifier was tested both against security reviews and against all reviews.

\subsubsection{Training and testing privacy classifier}
The pre-trained deep averaging network variation of USE was embedded as the input layer, followed by two hidden layers with dropout ratio 0.5 to reduce overfitting \cite{nema2022}, compiled with Keras' binary cross-entropy loss, with the best version across 20 epochs used for testing. On a test set of 100 reviews (50 security, 50 unrelated), the model achieved an accuracy of 0.79, recall of 0.81, and F1 score of 0.8; false positives arose from non-privacy issues resembling privacy complaints, false negatives from privacy issues framed around other concerns. Classifying every review extracted 20,622 privacy reviews, up from 6,187 via the similarity matrix method (0.038\% versus 0.13\% of the dataset), closely matching Nema, et al's finding of 0.15\% \cite{nema2022}.

\subsubsection{Training and testing permission classifier}
The same network structure was re-used for permission classification. Of the 20,622 security-related reviews, 4,638 correlated above 80\% with the permission GRs; these, along with 4,638 randomly chosen security reviews, were used for training. As shown in Table \ref{tab:classification}, classifying permission reviews from security reviews achieved an accuracy of 0.71, a recall of 0.79 and an F1 score of 0.75, while applying the model to the full review dataset achieved an accuracy of 0.82, a recall of 0.87 and an F1 score of 0.84.

\subsection{Data Analysis}
With a set of permission reviews identified, K-means clustering was used to surface recurring topics, which were then compared against app genre. The goal was to capture the sentiment and substance behind each review rather than the permission type mentioned, since two reviews about the same permission can convey very different attitudes. Since permission topics were not known in advance, an unsupervised K-means approach was used, with each cluster examined by manually reading individual reviews and identifying frequently occurring words to assign a meaningful topic label. The silhouette method was used to select an optimal k value, accounting for both intra-cluster similarity and inter-cluster dissimilarity; calculated for k = 2 through 10 (Figure~\ref{fig:silhouette}), the score reached a maximum at k = 7, which was used for clustering.

\section{Results}\label{results}
\subsection{RQ1: Can permission-related reviews be classified from a dataset of user reviews?}

From the data in Table~\ref{tab:classification}, classifying permission related reviews from the dataset of all reviews yielded the highest accuracy, followed by permission reviews from security reviews and finally security reviews from all reviews. Classifying permission reviews from all reviews proved easier than classifying them from security reviews alone, implying that the two categories share overlapping semantics and in some cases, near-identical phrasing, such as reviews following the form ``[permission/security related exclamation], this app is insecure, do not download!'' This structural similarity would make the boundary between them harder for the model to learn.
\vspace{-12pt}
\begin{table}[htbp!]
\caption{Security and permission classification results}
\label{tab:classification}
\centering
\small
\begin{tabular}{lccc}
\toprule
Classification & Accuracy & Recall & F1 score \\
\midrule
Security & 0.79 & 0.81 & 0.8 \\
Permission from all & 0.82 & 0.87 & 0.84 \\
Permission from security & 0.71 & 0.79 & 0.75 \\
\bottomrule
\end{tabular}
\end{table}
\vspace{-20pt}
The high accuracy but moderate recall observed in the all reviews classification reflects the model latching onto specific surface features of permission reviews, allowing it to quickly discard positive or non-security reviews that dominate the dataset. Using the model to select permission related reviews from the original HR data set returned 6,532 reviews making up 30\% of security reviews, nearly aligning with Nema, et al's research that found 50\% of security reviews are related to permissions \cite{nema2022}.

\begin{tcolorbox}[arc=0mm,width=1\columnwidth,
                  top=0mm,left=0mm,  right=0mm, bottom=0mm,
                  boxrule=.75pt]
\textbf{RQ1 Findings:} Permission-related reviews can be meaningfully classified from a dataset spanning all review types with an accuracy of 0.82.
\end{tcolorbox}

\subsection{RQ2: What topics do users discuss in permission-related reviews?}

\begin{table}[htbp!]
\caption{Permission Review Cluster Results}
\label{tab:clusters}
\centering
\small
\begin{tabularx}{\linewidth}{|c|p{2.2cm}|X|}
\hline
\textbf{Size} & \textbf{Identifier} & \textbf{Representative Review} \\
\hline
830 & Hostile & Horrible spying app. Needs my location? Uninstalling! \\
\hline
664 & Question & Why does this app need access to my contacts? \\
\hline
522 & Statement & App requests excessive personal information. \\
\hline
492 & Review & Privacy nightmare. Read the permissions carefully. \\
\hline
474 & Disappointment & Asked for passwords and credit card data. Uninstalled. \\
\hline
380 & Warning & Shares data without permission. Don't install. \\
\hline
142 & Removal & Google should remove this app for abusing permissions. \\
\hline
\end{tabularx}
\end{table}

Table~\ref{tab:clusters} summarises each cluster, from smallest to largest, with a sentiment and a representative review. Rather than aligning with specific permission types like location or camera access, clusters were shaped around the sentiment and attitude users expressed toward permissions in general.

\textbf{Hostile:} The largest cluster shows users threatening to ``uninstall immediately'', typically over permissions involving sensitive financial data, passwords, or PII, suggesting users are most vocal when data they consider deeply personal is requested.

\textbf{Question and Statement:} These clusters reflect users questioning why a permission is needed, or making neutral statements (e.g., ``This app keeps requesting access to my [data]''), both less aggressive than the Hostile cluster and often found in apps that seemed to request unnecessary data.

\textbf{Review:} Users here reference reading privacy policies or EULAs and urge others to do the same, having identified a permission they consider unjustified.

\textbf{Disappointment:} This cluster mixes positive reviews from users who liked the app but were put off by new permission requirements, often following an update; some stopped using the app, others simply expressed disappointment.

\textbf{Warning and Removal:} Warning reviews caution others against installing over alleged data misuse, while Removal reviews go further, asking Google to remove or flag the app, often citing ``fraud'' or ``theft''.

Each topic was compared against app genre (Table~\ref{tab:genre}, excluding genres with negligible permission-related reviews). The standout genres, finance, communication and productivity, are discussed further in Section \ref{dis}.
\vspace{-10pt}
\begin{table}[htbp!]
\caption{Comparing each topic to genre of review. Table shows percentage of each topic found per genre}
\label{tab:genre}
\centering
\footnotesize
\begin{tabular}{lccccccc}
\toprule
Genre & Host. & Quest. & Statem. & Review & Disapp. & Warn. & Removal \\
\midrule
Photography & 8 & 5 & 11 & 6 & 13 & 16 & 14 \\
Finance & 22 & 21 & 7 & 3 & 27 & 13 & 6 \\
Business & 7 & 3 & 2 & 2 & 5 & 4 & 1 \\
Social & 6 & 3 & 6 & 12 & 2 & 5 & 6 \\
Productivity & 5 & 20 & 16 & 9 & 8 & 5 & 17 \\
Communication & 23 & 11 & 12 & 41 & 13 & 23 & 14 \\
Tools & 0 & 6 & 7 & 4 & 1 & 4 & 4 \\
\bottomrule
\end{tabular}
\end{table}
\vspace{-20pt}
\begin{tcolorbox}[arc=0mm,width=1\columnwidth,
                  top=0mm,left=0mm,  right=0mm, bottom=0mm,
                  boxrule=.75pt]
\textbf{RQ2 Findings:} Topics in permission related reviews are shaped by user sentiment, not by the specific type of permission involved. The same underlying topics appear across all apps, but in different proportions depending on what users expect the app to do with their data.
\end{tcolorbox}

\section{Discussion}\label{dis}
Permission topic clusters reveal that users' discomfort with permissions is largely emotional: the hostile cluster being the biggest suggests developers who clearly explain permissions attract fewer negative reviews. Sentiment varies by genre: communication apps had nearly half of reviews in the Review cluster. Finance apps account for a fifth of hostile and questioning reviews and a quarter of disappointed reviews, often following an update introducing a new permission, while productivity apps saw more neutral questioning. The smallest clusters, Warning and Removal, suggest malicious behaviour is rare, and users are more often simply frustrated or confused. These findings have practical value: surfacing such reviews at installation could inform users, help developers clarify permissions, and let administrators flag dangerous apps at scale, as seen with similar monitoring strategies across other security-critical systems \cite{F5}.

\section{Limitations}\label{limit}
Limitations of this study fall in two main areas: the use of AI-generated data in the training pipeline and the size of test sets used for analysis. Although GRs were used to select training data rather than to train the classifier directly, it could be argued that all selected data was biased to the generated data; however, similarity matrices (Figure~\ref{fig:similarity}) show most reviews were no more than 50\% similar to each other, and HRs were selected on being at least 80\% similar to a GR, leaving up to 20\% free to vary, likely allowing the network to generalise beyond the matrix. Reproducibility is harder to guarantee since generated data was not used to train the model directly, though the method remains reproducible in principle. A related limitation is the test set size: limited to 100 each due to annotation effort, relatively small compared to similar studies, meaning a larger set would provide stronger validation.

\section{Conclusion}\label{conc}
As AI mobile apps become common, users are asked to share personal data with systems they may not fully understand, yet research shows most users have limited awareness of what they are consenting to. This paper asked whether user app reviews could identify permission-related concerns at scale, proposing a method to classify and cluster permission-related reviews to better understand user sentiment. The resulting classifier identified permission-related reviews with an accuracy above 82\%, revealing that user concerns organise not around specific permission types but around the sentiment users hold toward the requesting app, varying by app genre. Developers could draw on these reviews to identify which requests are generating concern, while administrators could use warning and removal clusters to flag apps misusing data access. Further research could build a more accurate, larger-scale classifier that produces detailed permission information without supervision.

\bibliographystyle{splncs04}
\bibliography{reference}

\end{document}